\documentclass[conference]{IEEEtran}
\IEEEoverridecommandlockouts
\usepackage{cite}
\usepackage{amsmath,amssymb,amsfonts}
\usepackage{algorithmic}
\usepackage{graphicx}
\usepackage{textcomp}
\usepackage{xcolor}
\usepackage[colorinlistoftodos]{todonotes}
\usepackage{hyperref}
\usepackage[most]{tcolorbox}
\usepackage{subcaption}
\usepackage{xargs}
\usepackage{makecell}
\usepackage[algo2e]{algorithm2e}
\usepackage{algorithm}
\usepackage{algorithmic}
\usepackage{multirow}
\usepackage{multicol}
\usepackage{float}
\usepackage{hhline}
\usepackage{orcidlink}

\newtheorem{example}{Example}
\newcommandx{\toDiscuss}[2][1=]{\todo[linecolor=red,backgroundcolor=red!25,bordercolor=red,#1]{#2}}
\newcommandx{\done}[2][1=]{\todo[linecolor=blue,backgroundcolor=blue!25,bordercolor=blue,#1]{#2}}

 \newcommand{\new}[1]{{{\textcolor{red}{\textsl{#1}}}}}

\newcommand{\set}[1]{\mathcal{#1}}
\newcommand{\domain}[1]{\mathbb{#1}}
\newcommand{\func}[1]{\texttt{#1}}

\newcommand{\edge}[2]{\overrightarrow{#1 #2}}

\def\BibTeX{{\rm B\kern-.05em{\sc i\kern-.025em b}\kern-.08em
    T\kern-.1667em\lower.7ex\hbox{E}\kern-.125emX}}
\begin{document}

\title{Prompting or Fine-tuning? A Comparative Study of Large Language Models for Taxonomy Construction 
 \thanks{\textsuperscript{*} Equal contribution}

 \thanks{Partially supported by the FRQNT-B2X project (file number: 319955) and the Wallenberg AI, Autonomous Systems
and Software Program (WASP), Sweden}
}

\author{\IEEEauthorblockN{Boqi Chen \textsuperscript{*}\orcidlink{0000-0002-1451-3603}}
\IEEEauthorblockA{\textit{Electrical and Computer Engineering} \\
\textit{McGill University}\\
Montreal, Canada}
\and
\IEEEauthorblockN{Fandi Yi \textsuperscript{*} \orcidlink{0009-0006-5373-2741}}
\IEEEauthorblockA{\textit{Desautels Faculty of Management} \\
\textit{McGill University}\\
Montreal, Canada}
\and
\IEEEauthorblockN{D\'aniel Varr\'o \orcidlink{0000-0002-8790-252X}}
\IEEEauthorblockA{\textit{Link\"oping University} \\
\textit{McGill University}\\
Link\"oping, Sweden / Montreal, Canada}
}

\maketitle

\begin{abstract}
Taxonomies represent hierarchical relations between entities, frequently applied in various software modeling and natural language processing (NLP) activities. They are typically subject to a set of structural constraints restricting their content.
However, manual taxonomy construction can be time-consuming, incomplete, and costly to maintain.
Recent studies of large language models (LLMs) have demonstrated that appropriate user inputs (called \emph{prompting}) can effectively guide LLMs, such as GPT-3, in diverse NLP tasks without explicit (re-)training. However, existing approaches for automated taxonomy construction typically involve \emph{fine-tuning} a language model by adjusting model parameters.  
In this paper, we present a general framework for taxonomy construction that takes into account structural constraints. We subsequently conduct a systematic comparison between the prompting and fine-tuning approaches performed on a hypernym taxonomy and a novel computer science taxonomy dataset. 
Our result reveals the following: (1) Even without explicit training on the dataset, the prompting approach outperforms fine-tuning-based approaches. Moreover, the performance gap between prompting and fine-tuning widens when the training dataset is small. However, (2) taxonomies generated by the fine-tuning approach can be easily post-processed to satisfy all the constraints, whereas handling violations of the taxonomies produced by the prompting approach can be challenging. These evaluation findings provide guidance on selecting the appropriate method for taxonomy construction and highlight potential enhancements for both approaches.

\end{abstract}

\begin{IEEEkeywords}
taxonomy construction, domain-specific constraints, large language models, few-shot learning, fine-tuning
\end{IEEEkeywords}

\section{Introduction}
\label{sec:introduction}

\textit{Context:}
Taxonomies are models that describe hierarchical relations between concepts or entities within a particular domain. The structure of a taxonomy is commonly limited by some taxonomy-specific constraints \cite{cocos2018comparing}. Hierarchical relations have significant importance in software engineering practice, including composition or generalization relations in domain modeling, hierarchical models in database systems, and inheritance in most object-oriented languages. Additionally, taxonomies are extensively applied in semantic web applications, typically in the form of OWL ontologies \cite{mcguinness2004owl}. 

Nevertheless, manually constructing taxonomies is typically time-consuming, prone to be incomplete, and challenging to maintain. For example, the relations in the WordNet taxonomy have been demonstrated to be incomplete  \cite{bansal2014structured}. To address the issues associated with manual construction, many approaches aim to automate taxonomy construction from a set of concepts \cite{bansal2014structured, mao2018end, Chen_2021}. However, these methods often focus on hypernym taxonomies utilizing widely available datasets, while their performance in other domains lacking such datasets remains less explored. Furthermore, although existing approaches incorporate structural constraints for taxonomies \cite{mao2018end, Chen_2021}, they rely on tree structures within the hypernym taxonomy and lack generalizability to other types of constraints.

\textit{Problem statement:}
The recent advances in generative large language models (LLMs) spark growing interest in their application to diverse software engineering activities, offering the potential for tackling classic software engineering challenges with novel approaches. Several powerful LLMs have demonstrated remarkable performance across various tasks relying solely on instructive input, without substantial task-specific training data. Consequently, an important question arises: in scenarios where some training data is available, is it still necessary to train a specialized model, or would instructive input prompts provided to a generative LLM be sufficient?

Our paper aims to address this question for automated taxonomy construction. Given a set of concepts with semantically meaningful names, the goal is to derive a taxonomy that captures the hierarchical relations among these concepts while fulfilling the \textit{structural constraints}.

\textit{Objective:}
This paper presents a comparative study using LLMs for taxonomy construction with datasets from two domains. Specifically, we compare two approaches: (1) a widely adapted approach that trains an LLM on a dedicated training set for taxonomy construction (referred to as ``\emph{fine-tuning}") and (2) a recent approach that provides instructive input to guide powerful generative LLMs (referred to as ``\emph{prompting}"). Our objective is to identify the strengths and limitations of both approaches and discuss potential improvements.


\textit{Contribution:}
Given a set of concepts, we propose a general framework to generate a taxonomy using LLMs. We subsequently compare two taxonomy construction methods within this general framework. The specific contributions of this paper can be summarized as follows.

\begin{itemize}
    \item We propose a general framework for taxonomy construction, which builds upon existing approaches to generate taxonomies from LLM-based relation predictions. 
    \item We formulate the taxonomy generation problem as a text generation task and propose a few-shot prompt for constructing taxonomies without explicit training. 
    \item We provide a new dataset specific to the computer science domain derived from the hierarchical relations within the ACM Computing Classification System (CCS).
    \item We conduct experiments using two fine-tuning approaches and a novel prompting approach for taxonomy construction, using two distinct taxonomy datasets. The quality of the generated taxonomies is evaluated against the ground truth, along with their structure consistency regarding the constraints.
\end{itemize}

\textit{Added value:}
To the best of our knowledge, this paper presents the first use of generative LLMs for taxonomy construction without explicit training. We present a comprehensive comparison of this approach against fine-tuning approaches. Furthermore, we extend the scope of our experiments beyond commonly adapted hypernym taxonomy by incorporating a novel domain-specific taxonomy focusing on computer science concepts. In our evaluation, we also assess both the performance of the generated taxonomies against the ground truth and their structure consistency. This paper provides valuable insights into leveraging LLMs for prediction hierarchical relations and demonstrates the impact of consistency in enhancing the performance of machine learning components. 


\section{Background}
\label{sec:background}

\subsection{Taxonomy Construction}
Given a set of concepts, taxonomy construction involves identifying the hierarchical relations between these concepts, such as parent-child or inclusion relations. Formally, a taxonomy is defined as a \textit{directed} graph $T=(\set{N}, \set{E})$, where $\set{N}$ represents the set of concepts (nodes) and $\set{E}$ represents hierarchical relations (edges) between concepts where $\edge{A}{B} \in \set{E}$ denotes that concept $A$ is a parent of concept $B$. Taxonomies are typically subject to specific structural constraints imposed by the domain \cite{cocos2018comparing}. 
Typically, a taxonomy is always \textit{connected} and \textit{acyclic}. Depending on whether a concept in the taxonomy can have multiple parents, the taxonomy may form a tree (for uni-parent) or DAG (for multi-parent). Relations in a taxonomy may also be transitive: if A is a parent of B, B is a parent of C, then A is also a parent of C.


A taxonomy is considered to \textit{consistent} if it satisfies these constraints. The constraints are denoted as $\set{C}$. The task of \textit{taxonomy construction} aims to create a taxonomy based on a set of concepts while aiming to satisfy these constraints \cite{bansal2014structured}. 

\begin{figure*}[tb]
    \centering
    \includegraphics[width=.9\linewidth]{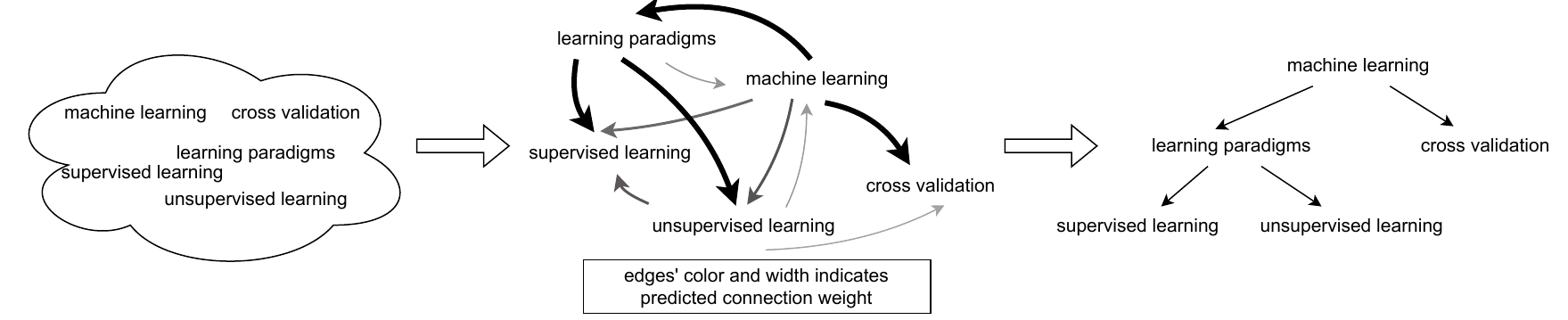}
    \caption{An example of taxonomy construction}
    \label{fig:taxo_example}
\end{figure*}

\begin{example}
    \autoref{fig:taxo_example} illustrates an example of taxonomy construction. Given the set of five terms on the left, taxonomy construction generates a hierarchical graph of these terms, as shown on the right. The edges in the graph represent hierarchical relations, e.g., ``learning paradigms" is a parent of ``supervised learning".
\end{example}

This paper focuses on the relation prediction-based taxonomy construction, as described by Chen et al. \cite{Chen_2021}. This setup involves the initial prediction of relations for each concept pair, followed by the construction of output taxonomy through a post-processing procedure. Formally, given a set of concepts denoted as $\set{N}$ and taxonomy constraints represented by $\set{C}$, the relation prediction process begins by assigning a score $s_{(A,B)}$ to each pair of concepts $A, B \in \set{N}, A \ne B$. The target is then to create a taxonomy $T=(\set{N}, \set{E})$ that maximizes the sum of the edges scores while adhering to the specified constraints. 


\subsection{Taxonomy Relations Applications}
Taxonomy relations exist in many knowledge engineering and modeling applications. The most common application in knowledge engineering is the \textit{hypernym} relations in lexicon databases such as WordNet \cite{miller1995wordnet}, while the taxonomy relations form a tree of related words. Inheritance (or generalization) is also a type of taxonomy relation and typically exists in most object-oriented languages, domain modeling, relational databases, and OWL ontologies. Depending on whether multi-parent is allowed, the relations can either form a tree of DAG. Finally, aggregation and composition relations in UML class diagrams can also be seen as tree-structured taxonomies.


\subsection{Language Models}
\paragraph{Large language models (LLMs)}
Language modeling is a classic task in natural language processing (NLP) that involves predicting the probability of a sequence of tokens. Large language models (LLMs) leverage deep neural networks, commonly using transformer blocks \cite{vaswani2017attention}. 

For a sequence of input tokens (referred to as a \textit{prompt}) $s = \{s_1, s_2,...,s_{k-1}\}$, LLMs estimate the probability of the next token $P(s_k|s_1,...,s_{k-1})$. LLMs can also be used for text generation through auto-regression, where the newly predicted token is appended to the token sequence to predict subsequent tokens. In this scenario, a search method such as beam search is typically utilized to generate the final output sequence \cite{brown2020language}. However, such search methods also introduce variability to the output of LLMs across different runs for the same input.

After pre-training an LLM for language modeling, it can be customized for other tasks by fine-tuning, which involves updating a subset of its parameters \cite{radford2018improving}. Recently, LLMs such as GPT-3 \cite{brown2020language
} and GPT-4 \cite{openai2023gpt4} show remarkable generalizability, as they can be adapted to other tasks by providing a few examples within the prompt, without parameter updates \cite{brown2020language}. 

This paper explores the application of LLM fine-tuning and prompting techniques in the context of taxonomy construction. In particular, we compare fine-tuning approaches for widely-used open-source and cost-effective LLMs, while also investigating the potential of prompting techniques for more powerful, black-box LLMs accessible through an API. 


\paragraph{Fine-tuninig}
Fine-tuning of an LLM involves initially \textit{pre-training} the LLM with a large corpus for language modeling and subsequently adapting it to a different task by training a subset of the LLM's parameters using a task-specific dataset. In this paper, we explore two widely-used fine-tuning methods for LLMs: the \textit{layer-wise} approach and \textit{LoRA}.

\textbf{Layer-wise approach} is a classic fine-tuning method that focuses on updating only a subset of parameters in an LLM. For a pre-trained autoregressive LLM with $l$ layers, parameters from the last $z$ layers are updated and the initial $l-z$ layers are fixed during training for the task. This approach leverages the fixed layers of the LLM to capture task-independent high-level information from the input. Subsequently, this information is used by the fine-tuned layers to create task-specific output. 

\textbf{LoRA} (low rank adaptation) for LLMs is a fine-tuning approach that significantly reduces the number of parameters requiring updates in training \cite{hu2022lora}. This method treats the fine-tuning of each weight matrix $W \in \domain{R}^{(m, n)}$ as the learning of a delta matrix $\Delta W$ enabling the representation fine-tuned matrix $W'$ as the sum of the two matrices: $W'=W+\Delta W$. In many cases, $\Delta W$ exhibits low-rank characteristics 
\cite{li2018measuring,aghajanyan2020intrinsic}, indicating that it can be decomposed as the product of two smaller matrices $\Delta W = A \times B$, such that $A \in \domain{R}^{(m, d)}, B \in \domain{R}^{(d, n)}$. Typically, the value of $d$ is chosen so that the parameter count of matrices $A$ and $B$ is notably smaller than that of matrix $W$: $m \times d + d \times n \ll m \times n$. The adoption of LoRA can lead to significant reductions in memory overhead and storage costs during training, making it suitable for taxonomy construction that requires 
different models across different domains.

\paragraph{Prompting}
Instead of updating the weights of the LLM, in case of prompting, users provide instructions and examples as input of the LLM in order to guide its output towards the desired task. We focus on one popular prompting method known as \textit{few-shot prompting}.

\textbf{Few-shot prompting} aims to teach the LLM how to adapt to a specific task by carefully constructing an instruction that includes a few examples extracted from the task dataset. For example, in taxonomy construction, one can begin by providing instruction on how the taxonomy should be generated, followed by a list of example taxonomies, and concluding with the set of concept names as input to the taxonomy construction. It has been demonstrated that few-shot prompting can yield remarkable results, particularly when implemented with large-scale LLMs consisting of hundreds to thousands of billions of parameters \cite{brown2020language}. However, the effectiveness of this approach diminishes when applied to less powerful LLMs.



\section{Approach}
\label{sec:method}

\begin{figure}[tb]
    \centering
    \includegraphics[width=\linewidth]{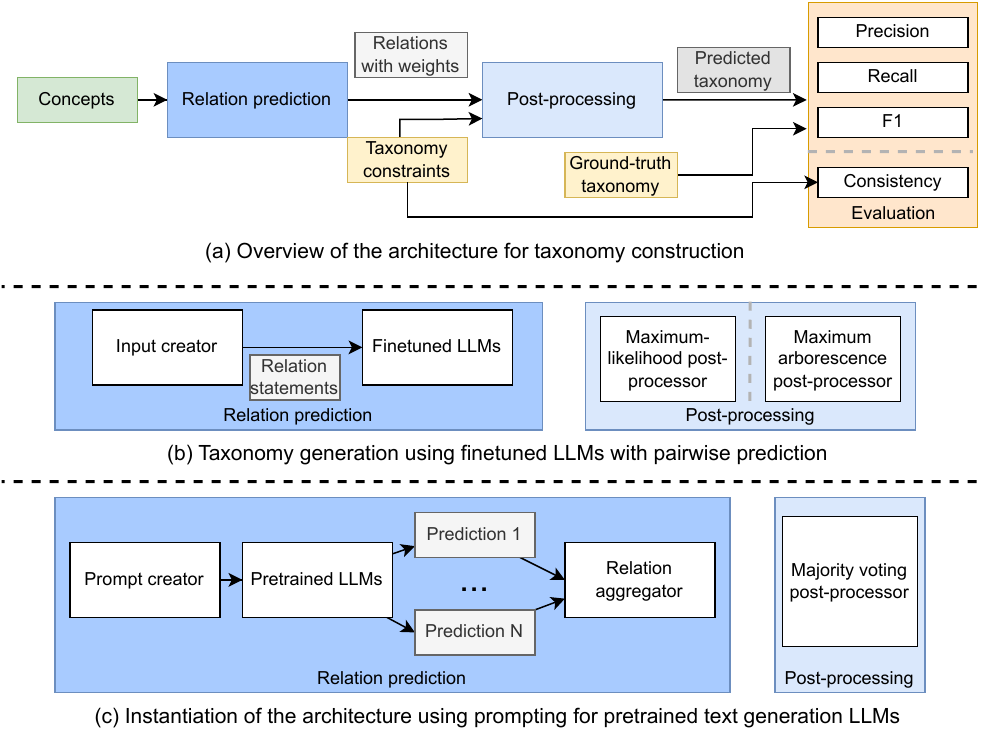}
    \caption{Architecture for taxonomy generation using LLMs. Common components are omitted for (b) and (c).}
    \label{fig:architecture}
\end{figure}

\subsection{Overview}
\paragraph{Problem formulation}
Given a set of concepts and constraints, \textit{taxonomy construction} is to create a taxonomy that accurately represents the hierarchical structure among these concepts based on their names. Formally, let $\set{N}$ denote a set of concepts, $\set{C}$ represent the constraints for the taxonomy, and $T$ be the ground-truth taxonomy of these concepts. The objective of taxonomy construction is to establish a mapping function $\func{f}$ such that the generated taxonomy $T'$ is obtained as $T'=\func{f}(\set{N}, {\set{C}})$. The generated taxonomy $T'$ should be consistent with the constraints in $\set{C}$ and \textit{similar} to the ground truth taxonomy $T$. Ideally, they should be identical: $T' \cong T$. In this paper, our objective is to evaluate the capability of LLMs in the context of $\func{f}$, considering both the fine-tuning and prompting approaches.

The structural constraints governing a graph can often be expressed using first-order logic or high-level constraint languages \cite{chen2022consistent}. For the sake of conciseness, we provide a textual description of these constraints. In this paper, our focus is on taxonomies that have a tree structure \new{\cite{cocos2018comparing}}. Therefore, we define the set of constraints $\set{C}$ as follows.

\begin{enumerate}
    \item \textbf{Uniqueness of the root}: There exists exactly one node that does not have a parent node in each taxonomy.
    \item \textbf{Uniqueness of the parent}: For all non-root nodes, exactly one parent node exists.
\end{enumerate}

\paragraph{Architecture}
The general architecture of our taxonomy generation approach is illustrated in the top diagram of \autoref{fig:architecture}. The construction of a taxonomy involves two main steps. Firstly, the \textit{relation prediction} component generates a set of candidate relations along with their associated weights. These weighted relations are then subjected to a \textit{post-processing} stage, which can take into consideration the imposed constraints to generate the final taxonomy. Finally, the predicted taxonomies are evaluated using the ground-truth taxonomy and metrics, which will be described in \autoref{sec:experiments}.


In the remainder of this section, we first discuss the general architecture in detail. Next, we explore two LLM-based approaches using the general architecture: fine-tuning (\autoref{fig:architecture}(b)) and prompting (\autoref{fig:architecture}(c)). These two approaches mainly differ in the generation of candidate relations and the post-processing of these relations. 



\subsection{General architecture}
\paragraph{Relation prediction}
The \textit{relation prediction} component aims to generate a set of candidate relations and their corresponding weights based on \textit{input concepts}. This component typically relies on a machine learning model, such as an LLM, to perform the prediction. Formally, let $\set{N}$ denote the set of input concepts. The relation prediction component produces a set of candidate relations $\set{E}$ along with a weight mapping $w: \set{E} \to \domain{R}$, where $w_{\edge{A}{B}}$ represents the weight assigned to the relation indicating that concept A is the parent of concept B. 

\paragraph{Post-processing}
Given a set of weighted edges, the post-processing stage aims to determine the most likely taxonomy based on these edges. Additionally, structure constraints may be taken into consideration during this process. Let $G=(\set{N}, \set{E})$ represent the weighted output graph of the relations predicted by the \textit{relation prediction} component and let $w$ be the corresponding weight mapping. Furthermore, let $w_{max}$ denote the maximum possible weight that an edge can have. For all edges $\edge{A}{B} \in \set{E}$, we define the \textit{weight of a subgraph} $S=(\set{N}, \set{E}_S)$ of $G$ as

\begin{equation}
    \label{equ:objective}
    \small{
    W(S, G) = \sum_{\edge{A}{B}\in \set{E}_S} w_{\edge{A}{B}} + \sum_{\edge{A}{B} \not \in \set{E}_S} (w_{max} - w_{\edge{A}{B}})
    }
\end{equation}

Then, the task of identifying the most likely taxonomy can be formulated as a \textit{constraint optimization} problem, aiming to find the \textit{subgraph with the maximum weight} while adhering to all the specified \textit{constraints}.

\begin{equation}
\label{equ:optim}
\small{
\begin{aligned}
    &\text{max}_{S} W(S, G) \:\:\text{subject to } c \in \set{C} 
\end{aligned}
}
\end{equation}
In this paper, we investigate both the above optimization problem, which considers the constraints, and a relaxed version of the problem that disregards these constraints.

\subsection{Fine-tuning LLMs}
Following fine-tuning for taxonomy construction, the LLM first assigns scores to \textit{all pairs} of input concepts. These scores represent the likelihood of hierarchical relations among these pairs. Then, a taxonomy is derived based on the relation scores. The overview of this approach is illustrated in \autoref{fig:architecture}(b).

\paragraph{Relation prediction}
For the fine-tuning of LLMs, we adopt the pairwise classification commonly used by others \cite{Chen_2021}

Given a concepts pair $(A, B)$, we frame the task as a sentence classification problem aiming to assign a label to a sentence. We use the LLM to classify a sentence expressing the relation between concept $A$ and $B$ as true or false. We use the following template to create the input to the LLM: \textit{I am doing the taxonomy research. I think $B$ is a subtopic of $A$}. For example, in \autoref{fig:taxo_example}, the input for \textit{machine learning} and \textit{supervised learning} would be expressed as \textit{I am doing the taxonomy research. I think supervised learning is a subtopic of machine learning}. Although various templates are feasible, previous research indicates they have minimal impact on performance \cite{Chen_2021}.
 
To construct a training set, we generate a collection of positive and negative examples based on a given ground-truth taxonomy. Considering that the number of negative examples can grow exponentially with the number of concepts, we employ \textit{negative sampling} \cite{mikolov2013distributed} to create negative examples.

\begin{enumerate}
    \item For each edge $\edge{A}{B}$ in the taxonomy, we use the template to generate a positive example.
    \item  We randomly select $N$ concept pairs $(A, B)$  where $B$ is not a descendant of $A$ in the taxonomy. 
\end{enumerate}
Then, the LLM is fine-tuned to predict the correctness of the input sentence using the cross-entropy loss. 

During inference, the \textit{relation prediction} component assigns scores to \textit{all pairs} of input concept as weighted relations. 


\paragraph{Post-processing} 
Given a set of concepts for inference, we first use the fine-tuned LLM to compute scores for \textit{all} concept pairs and subsequently identify the \textit{most likely} taxonomy based on these pairwise scores. In this scenario, we explore two distinct variations of the post-processor that correspond to different forms of the optimization problem described in \autoref{equ:optim}: the \textit{maximum-likelihood post-processor}, which disregards the constraints and the \textit{maximum arborescence post-processor}, which incorporated the constraints. 

\textbf{Maximum-likelihood:} 
The LLM is trained to predict pairwise relations between concepts while disregarding taxonomy constraints. The output of the LLM for all pairs of concepts can be interpreted as the weighted graph $G$, where the edge weights represent the logarithm of probability scores predicted by the LLM. Let $P(A, B)$ denote the LLM predicted probability score for concepts $A$ and $B$. We define the weight of each edge $w_{\edge{A}{B}}$ as $w_{\edge{A}{B}} = log(P(A, B))$
    
The weight of a subgraph in \autoref{equ:objective} is determined by the sum of the logarithmic probabilities of edges in the subgraph using the Naive Bayes assumption. When disregarding the constraints, the subgraph with the maximum weight can be obtained using the maximum likelihood approximation. For each edge $\edge{A}{B} \in \set{E}$, include $\edge{A}{B}$ to $S$ if $P(\edge{A}{B}) > 0.5$.



\textbf{Maximum arborescence:}
Disregarding all constraints as the maximum likelihood approach can create inconsistent taxonomies that affect performance. An approach to address the optimization problem in \autoref{equ:optim} involves using constraint optimization techniques, such as formulating the problem as MAXSAT and employing a constraint solver \cite{chen2022consistent}. However, this approach can be computationally expensive and ignores the graph structure inherent in the taxonomy construction task.  

Instead, we use the approach by Chen et al. \cite{Chen_2021}. The problem of optimizing the subgraph that adheres to all taxonomy constraints can be treated as a \textit{maximum spanning arborescence} problem. A spanning arborescence is analogous to a spanning tree in directed graphs. We use the Edmonds algorithm \cite{edmonds1967optimum} to identify the arborescence as the final output taxonomy, satisfying all imposed constraints for a tree-structured taxonomy.

\subsection{Prompting with GPT-3.5}
When using prompting with recent powerful LLMs such as GPT-3, accessed through APIs, the cost is typically determined by the API call count or tokens utilized in the request. Consequently, assigning scores to all pairs becomes highly expensive due to the quadratic growth in the number of pairs and, thus, a quadratic number of LLM calls. Alternatively, we introduce a more challenging task for the prompting setting, aiming to generate \textit{all} candidate relations directly from the set of concepts with one LLM call. This is formulated as a \textit{text generation} problem from concepts to relations. \autoref{fig:architecture}(c) shows the overview of the prompting approach.

\begin{figure}
    \centering
    \includegraphics[width=.85\linewidth]{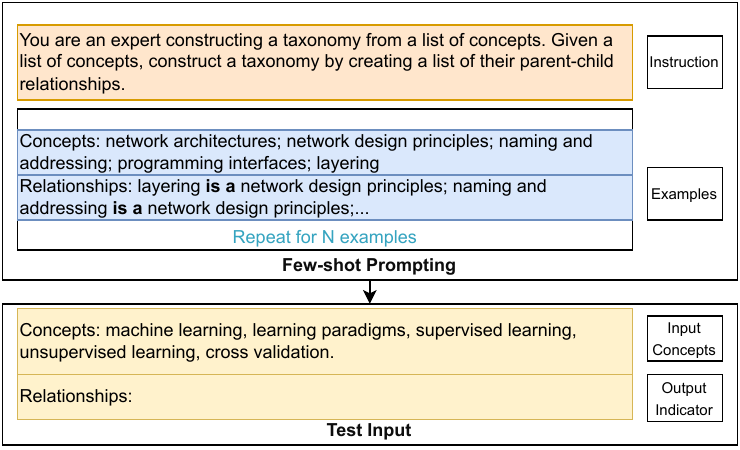}
    \caption{An example of the few-shot prompt}
    \label{fig:prompt}
\end{figure}

\paragraph{Relation prediction}
To adapt the generative LLM to the taxonomy construction task, we use a few-shot prompt consisting of examples \textit{randomly sampled} from the training set. An example of the few-shot prompt used in taxonomy construction is illustrated in \autoref{fig:prompt}. The prompt provides the instruction for the taxonomy construction task, followed by a selection of concept-relation pairs (randomly sampled) from the training dataset. To mitigate any potential bias in the LLM regarding specific ordering, we randomly shuffle the order of concepts and relations in the examples. Finally, the prompt presents the input concepts in a shuffled order. 

\paragraph{Post-processing}
Due to the inherent randomness in beam search used by generative LLMs \cite{brown2020language} and shuffle in the few-shot prompt, the outcome for generating taxonomy for the same concepts may vary across different runs. To address the issue of randomness, we propose a post-processing method that aggregates results from multiple runs through \textit{majority voting}, motivated by the self-consistency prompting technique \cite{wang2022self}. We disregard the taxonomy constraints in the prompting approach, deferring them to future works.

Using different few-shot examples, we obtain relation predictions for the same set of concepts $N$ times. Then, the \textit{relation aggregator} assigns a weight to each relation candidate based on their frequency of occurrence across the $N$ generations. Denoting the count of occurrences of relation $\edge{A}{B}$ in the $N$ generations as $c^N_{\edge{A}{B}}$, the weight of the relation is defined: $w_{\edge{A}{B}} = c^N_{\edge{A}{B}}$. Finally, the \textit{majority voting post-processor} includes a relation into the final taxonomy if it appears in more than half of the generations: $w_{\edge{A}{B}} \ge \lceil \frac{N}{2} \rceil$.



\section{Computer science Taxonomy dataset}
\label{sec:preprocessing}

\begin{figure*}[tb]
    \centering
    \includegraphics[width=\linewidth]{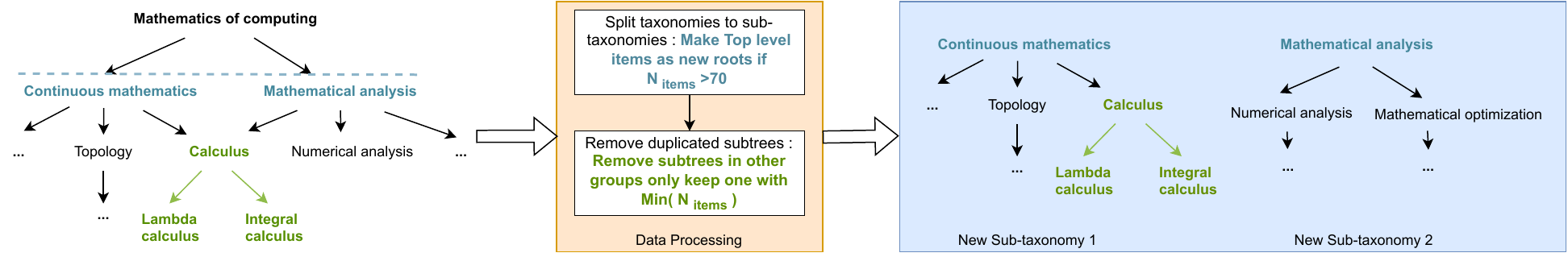}
    \caption{Preprocessing of the ACM CCS dataset}
    \label{fig:data_preprocessing}
\end{figure*}

For this paper, we create a novel taxonomy construction dataset based on the Association for Computing Machinery (ACM) Computing Classification System (CCS) \cite{ACM}, an established ontology in the field of computer science.

CCS is a domain-specific ontology that includes the hierarchical relations for various terms from computing fields. We first convert the ontology in OWL format into a pairwise dataset. There are 13 taxonomies, each containing a number of terms ranging in $[19, 323]$. As shown in \autoref{fig:data_preprocessing}, we apply a two-step data preprocessing method to transform the dataset into the standard format used in taxonomy construction \cite{bansal2014structured}. 

\textbf{Taxonomy split:}
The distribution of the number of nodes in the original taxonomies is skewed. The largest taxonomy consists of 323 concepts, which is 15 times larger than the smallest taxonomy, leading to a potential imbalance in the training process. To address this imbalance, following a similar approach to taxonomy processing in other domains \cite{bansal2014structured}, we split 13 taxonomies into 75 sub-taxonomies to achieve a more balanced distribution of terms. We split a taxonomy into sub-taxonomies if it has more than 70 terms. Specifically, the taxonomy is divided from the first level below its root (as depicted on the left side of \autoref{fig:data_preprocessing}), resulting in all the direct children of the root becoming new roots in the sub-taxonomies.

\textbf{Duplication removal:}
Following the split, we discover 144 duplicated relationships across the taxonomies. To prevent data leakage during training and achieve a more balanced distribution of the terms per taxonomy, we remove duplicates based on the following procedure. 
\begin{itemize}
    \item If some relations appear in multiple taxonomies, We keep only one instance of the duplicated relations in the taxonomy with the fewest terms among those taxonomies.
     \item However, if all relations in a taxonomy appear in other taxonomies, we remove that taxonomy from the dataset.
\end{itemize}


\section{Experiments}
\label{sec:experiments}
In this section, we describe our experiment setup and evaluate the performance of the approaches. We compare two fine-tuning approaches with a few-shot prompting approach for LLMs. We evaluate the performance of these approaches in comparison to the ground truth and assess their consistency using the constraints. Furthermore, we discuss the implications of the experimental findings. Overall, the objective of this paper is to address the following research questions (RQs):
\begin{enumerate}
    \item How do the two LLM-based approaches differ when compared to the ground truth?
    \item What are the differences between the two LLM-based approaches in generating consistent taxonomies?  
\end{enumerate}

The implementation of approaches and the experiment setup can be found in the GitHub repository for this paper\footnote{\url{https://github.com/20001LastOrder/Taxonomy-GPT}}.

\subsection{Setup}

\paragraph{Dataset}
We evaluate the approaches using two taxonomy datasets representing English language concepts and computer science concepts, respectively. Each dataset is divided into training ($70\%$), validation ($15\%$), and testing ($15\%$), following the same splits as previous work \cite{bansal2014structured, Chen_2021}.


\textbf{WordNet} \cite{miller1995wordnet} is a hypernym taxonomy. We use a medium-sized variant of WordNet \cite{bansal2014structured}, which splits the original large taxonomy into several distinct subtrees. The dataset consists of 761 bottom-out non-overlapping taxonomies, containing 11 to 50 terms with a maximum tree height of three.
The dataset contains 14,477 unique terms with 14,877 positive pairs. 

\textbf{ACM CCS} is the novel taxonomy construction dataset from the ACM Computing Classification System as shown in \autoref{sec:preprocessing}. After the split, the dataset contains $75$ non-overlapping taxonomies.
While some taxonomies in the dataset contain nodes with multiple parents, deviating from the tree structure, we still apply the maximum arborescence post-processor for this dataset and leave the handling of such structures as future work. 

\paragraph{Compared approaches}
We compare two taxonomy construction approaches based on LLMs: fine-tuning and prompting, with multiple settings for each approach. 

\textbf{Fine-tuning:}
We perform fine-tuning on a GPT-NEO model \cite{gpt-neo} with 1.3 billion pre-trained parameters for language modeling on the Pile dataset \cite{gao2020pile}. Since we focus on comparing finetuning and prompting methods, to minimize the impact of other factors such as model architecture, we omit masked language models like BERT with different model architectures than GPTs. We use two fine-tuning techniques for relation prediction: (1) \textbf{layer-wise}, where only parameters in the last few layers are updated during fine-tuning, and (2) \textbf{LoRA}, which updates all model parameters using low-rank decomposition. In the layer-wise approach, the parameters in the last five linear layers of the transformer block and the final linear layer are updated during fine-tuning. For LoRA, we set $d=8$ and apply a LoRA dropout rate of 0.1.

In this setting, we use two post-processing approaches: (1) \textbf{MALI}: which is the \emph{maximum likelihood} post-processor that selects the most probable subgraph from the relation predictions, and (2) \textbf{MSA}, which considers the taxonomy constraints and determines the \emph{maximum spanning arborescence}. 

\textbf{Prompting:}
We access the \textbf{GPT-3.5-turbo} model through the OpenAI API\footnote{\url{https://platform.openai.com/docs/models/gpt-3-5}}. Specifically, we choose the snapshot of this model from June 13th, 2023. We select this model due to its balanced trade-off between inference performance and cost, making it suitable for comparison with the fine-tuning approaches. While more powerful models such as GPT-4 exist, their cost is over 20 higher than the GPT-3.5 models\footnote{\url{https://openai.com/pricing}}, making them considerably more expensive than fine-tuning an LLM. We refer to this model as \textbf{GPT-3.5} throughout the experiment.

We conduct five generations using GPT-3.5 to generate candidate relations, aiming to mitigate the impact of randomness. For the WordNet dataset, we randomly select five taxonomies from the training dataset as few-shot examples. Since taxonomies from the CCS dataset are typically larger, we randomly select three as few-shot examples. Then, we compare two setups: (1) \textbf{IB}: The \emph{individual best} performance among the five runs based on the evaluating metric and (2) \textbf{MV}: The taxonomy obtained through the \emph{majority voting} post-processor. 

\subsection{RQ1 Taxonomy quality}
\paragraph{Setup}
This research question focuses on evaluating the generated taxonomies from different approaches in comparison to the ground truth taxonomies. Additionally, we include the state-of-the-art performance achieved on the WordNet dataset (\textbf{SOTA}). Since the CCS dataset is newly constructed, we establish a baseline by utilizing the performance of a randomly initialized LLM with the maximum arborescence post-processor (\textbf{Random}). Following prior research \cite{bansal2014structured, mao2018end, Chen_2021}, we compare the generated and ground truth taxonomies by examining the ancestral relations. Concept A is considered \textit{an ancestor} of concept B if there exists a path from A to B. We report the average precision, recall, and the $F_1$ score of the ancestral relations among all taxonomies. 


\begin{table*}[tb]
    \footnotesize{
    \centering
    \begin{subtable}[h]{0.45\textwidth}
    \centering
    \begin{tabular}{|l|c|c|c|c|}
    \hline
    Settings &  Precision (\%) & Recall (\%) & F1 (\%)\\
    \hline
    Layer-wise (MSA)    & 62.91 & 50.07 &  54.58 \\
     \hline
     Layer-wise (MALI)    & 67.66 & 39.08 &  42.29 \\
     \hline
     LoRA (MSA) & 62.57 & 51.21 & 55.03 \\
     \hline
     LoRA (MALI) & 66.98 & 42.73 & 46.97 \\
    \hhline{|====|}
     GPT-3.5 (IB) & 64.64 & \textbf{53.65}  & 57.25\\
     \hline
     GPT-3.5 (MV) & \textbf{77.50} & 47.98 & \textbf{57.26} \\
     \hhline{|====|}
     SOTA \cite{Chen_2021} & 67.30 & 62.00 & 63.50 \\
     \hline
    \end{tabular}
    \label{tab:wordnet_result}
    \end{subtable}
    \hfill
    \begin{subtable}[h]{0.45\textwidth}
    \centering
    \begin{tabular}{|l|c|c|c|c|}
    \hline
    Settings &  Precision (\%) & Recall (\%) & F1 (\%) \\
    \hline
    Layer-wise (MSA)    & 43.36 & 41.83  &  41.65  \\
     \hline
     Layer-wise (MALI)    & 26.23 & 55.46 &  26.51  \\
     \hline
     LoRA (MSA) & 44.18 & 45.45 & 43.60  \\
     \hline
     LoRA (MALI) & 33.16  & 51.68 & 33.46   \\
    \hhline{|====|}
     GPT-3.5 (IB) & 66.23 & \textbf{63.18}  & 61.07 \\
     \hline
     GPT-3.5 (MV) & \textbf{84.73} & 55.80 &  \textbf{64.18} \\
     \hhline{|====|}
     Random  & 16.21 & 17.44 &  16.07 \\
     \hline
    \end{tabular}
    \label{tab:ccs_result}
    \end{subtable}
    }
    \caption{Evaluation results for the WordNet (left) and CCS (right) dataset, averaged over all test taxonomies}
    \label{tab:performance_result}
\end{table*}

\paragraph{Results}
\autoref{tab:performance_result} shows the performance of all settings compared to the ground truth on the two datasets. We group the settings by the approach of relation prediction. 

Overall, incorporating constraint-aware post-processor enhances the quality of the generated taxonomies by improving consistency. In the case of WordNet, while there is a marginal decrease in precision (approximately 5\%) when comparing MSA and MALI across both layer-wise and LoRA settings, MSA has substantial improvement in recall and $F_1$ (around $10\%$) compared with MALI. Interestingly, there is a slight decrease in recall for MSA in case of CCS while improving in the precision and $F_1$ scores. This difference can be attributed to the domain and size distinctions between the two datasets. Similarly, the MV setting for GPT-3.5 also enhances precision and $F_1$ score for both datasets while sacrificing recall, compared with the best result for individual generation. 

In comparison to the fine-tuning settings (layer-wise and LoRA), the prompting approach (GPT-3.5) yields better taxonomies. When considering the $F_1$ score, the best prompting setting (MV) outperforms the best fine-tuning setting on WordNet by $2.23\%$. In CCS, the difference becomes more significant, exceeding $20\%$. We attribute this gap in performance improvement to the smaller size of the CCS dataset, which makes the fine-tuning of an LLM more challenging. 


Furthermore, even the best setting of GPT-3.5 falls short of the current state-of-the-art performance by $6.24\%$ in $F_1$, suggesting further enhancements through prompting and post-processing techniques. 
On the CCS dataset, all fine-tuning and prompting settings outperform the random baseline.


\begin{tcolorbox}
\textbf{Answer to RQ1.}  
The prompting method outperforms the fine-tuning method in both datasets when comparing the $F_1$ score. Moreover, the gap between the two methods increases when the training dataset is smaller.
\end{tcolorbox}

\subsection{RQ2 Taxonomy Consistency} 
\paragraph{Setup}
This section focuses on quantifying the consistency of generated taxonomies. To assess the inconsistency level resulting from violations, we use the following four metrics. 
(1) \textbf{\#Roots} represents the average number of roots per generated taxonomy. The ideal value should be one.
(2) \textbf{NRG (\%)} (No root graph) is the percentage of taxonomies without root. This scenario may arise due to cycles in the taxonomy, and a consistent taxonomy must always include a root.
(3)\textbf{\#Parents} indicates the average number of parents of all non-root nodes across the taxonomies. In a consistent taxonomy, each non-root node should have precisely one parent. 
(4) \textbf{MPN (\%)} (Multi-parent nodes) is the average percentage of non-root nodes with multiple parents across the taxonomies. In a consistent taxonomy, the value for this metric should be 0\%. 
The first two metrics correspond to the \textbf{uniqueness of the root} constraint, and the latter two metrics are for the \textbf{uniqueness of the parent} constraint.

\paragraph{Results}
As shown in \autoref{tab:violation_res}, all taxonomies generated by the fine-tuning methods (layer-wise and LoRA) with MSA are consistent with the imposed constraints. The MALI setting performs poorly for NRG, indicating the presence of cycles without root and leaf in certain taxonomies.
Regarding the prompting method (GPT-3.5), IB and MV settings yield inconsistent taxonomies. However, the extent of inconsistency is much less compared to the MALI setting in fine-tuning.

When comparing the WordNet and CCS datasets, the taxonomies generated in the CCS dataset are more consistent than those in WordNet. These differences may originate from the difference in dataset domains. The CCS dataset specifically contains computer science terms, whereas WordNet represents a hypernym taxonomy with more general terms. 


\begin{tcolorbox}
\textbf{Answer to RQ2.}  
The fine-tuning methods can produce taxonomies satisfying all constraints with the MSA post-processor. However, taxonomies generated by the prompting approaches still violate some constraints.

\end{tcolorbox}

\begin{table*}[tb]
    \centering
    \begin{subtable}[h]{0.45\textwidth}
    \centering
    \begin{tabular}{|l|c|c|c|c|}
    \hline
    Approaches & \#Roots& NRG (\%) & \#Parents &MPN (\%)\\
    \hline
    Layer-wise (MSA)  &  1.00&0.00&1.00&0.00\\
     \hline
     Layer-wise (MALI)  &1.48&16.51&1.47&31.28\\
     \hline
     LoRA (MSA) & 1.00&0.00&1.00&0.00\\
     \hline
     LoRA (MALI) & 1.06&43.36&2.56&66.32\\
    \hhline{|=====|}
     GPT-3.5 (IB) & 2.00&4.39& 1.00&0.04\\
     \hline
     GPT-3.5 (MV) &2.25&1.75&1.00&0.15\\
     \hline
    \end{tabular}
    \label{tab:viol_wordnet_result}
    \end{subtable}
    \hfill
    \begin{subtable}[h]{0.45\textwidth}
    \centering
    \begin{tabular}{|l|c|c|c|c|}
    \hline
        Approaches & \#Roots& NRG (\%) & \#Parents &MPN (\%)\\
    \hline
    Layer-wise (MSA)  &  1.00&0.00&1.00&0.00\\
     \hline
     Layer-wise (MALI)  &0.57&64.29&3.30&59.81\\
     \hline
     LoRA (MSA) & 1.00&0.00&1.00&0.00\\
     \hline
     LoRA (MALI) & 1.36&28.57&2.60&65.13\\
    \hhline{|=====|}
     GPT-3.5 (IB) &1.27&0.00&1.00&0.00\\
     \hline
     GPT-3.5 (MV) &2.87&0.00&1.01&0.78\\
     \hline
    \end{tabular}
    \label{tab:viol_ccs_result}
    \end{subtable}
    \caption{Consistency evaluation for the WordNet (left) and CCS (right) dataset, averaged over all test taxonomies}
    \label{tab:violation_res}
\end{table*}

\subsection{Discussion}
This section explores various implications of the experimental findings and discusses the limitations of the experiments.

\textbf{Approach selection:}
In RQ1, we observe that prompting a powerful LLM outperforms fine-tuning a weaker LLM. This performance gap becomes more significant when the training dataset is smaller. The scarcity of data is one factor hindering the adaptation of machine learning-based approaches to software engineering challenges. Our experimental finding demonstrates the potential of leveraging powerful LLMs through prompting techniques in addressing such challenges. We also note none of the approaches beat the current \textbf{SOTA} method based on BERT in $F_1$ score on WordNet. This suggests the model architecture may impact the task performance. The task may benefit from the bi-directional information flow in BERT compared to the left-to-right flow in GPTs.

However, Our experiments focus on two domains with publicly accessible datasets, which may be used as a part of the training data for the powerful LLM, resulting in good performance during evaluation. Besides, the setup assumes that concepts have semantically meaningful names, which might not be applicable to scenarios where concepts have ambiguous names or names heavily reliant on contextual interpretation. Future work will focus on overcoming these limitations.



\textbf{Taxonomy consistency:}
In RQ1, we find that enhancing consistency through a constraint-aware post-processor improves the quality of the generated taxonomy in the fine-tuning approach. However, in RQ2, we discover that the majority voting post-processor of the prompting method still generates inconsistent taxonomies. This suggests one possible way to further improve the prompting performance by developing a constraint-aware post-processor tailored to this approach.

We focus on two constraints that enforce the tree structure of taxonomy. However, some ground truth taxonomies in the CCS dataset violate this condition by having nodes with multiple parents. Additionally, other domains may impose other types of constraints on taxonomies. Therefore, one can explore methods for generating consistent taxonomies from both approaches while accounting for general constraints.

Additionally, the behavior of online LLMs, such as ChatGPT, may change due to internal updates of the model \cite{chen2023chatgpt}. To avoid this change, we use the OpenAI API and choose a fixed model snapshot. Our approaches do not distinguish the definition of relations in the taxonomy. For example, while hypernym relations in WordNet and domain-specific relations in CCS both represent hierarchical relations, their definitions differ. One may encode this information to further improve the performance. We will leave this improvement as future work.


\section{Related Work}
\label{sec:rel_work}
\textbf{Automated taxonomy construction:}
Most existing approaches use NLP-based methods  automatically constructing hypernym taxonomies, such as WordNet. Bansal et. al. \cite{bansal2014structured} first discuss the task of taxonomy construction using WordNet sub-trees and propose a method based on belief propagation. Mao et. al. \cite{mao2018end} propose an end-to-end approach for taxonomy construction using reinforcement learning. Chen et. al. \cite{Chen_2021} present a taxonomy construction method by training a masked language model and outperforming previous methods. Recent studies have shown a growing interest in taxonomy construction using minimally supervised learning through the adaptation of pre-trained language models \cite{jain2022distilling, sun2022minimally}.

\textbf{LLMs for MDE:}
Despite the growing interest of LLMs such as GPTs \cite{brown2020language} and BERT \cite{devlin2018bert}, their application in model-based software engineering (MDE) remains limited and has yet to achieve widespread adoption. Due to the sparsity of data, existing approaches typically involve fine-tuning using a small dataset, leading to modeling tools such as recommender systems for meta-modeling \cite{weyssow2022recommending}. Still, the emergence of powerful generative LLMs like GPT4 \cite{openai2023gpt4} paves the way for novel methods in adapting LLMs in MDE \cite{combemale2023chatgpt}. Specifically, there is a growing interest in using LLMs with prompting techniques. These approaches are often in an interactive context, such as using GPT-3 for model completion\cite{chaaben2022towards} or creating a chat-like experience with ChatGPT for UML modeling \cite{camara2023assessment}

Compared to previous studies, we provide a systematic analysis of LLMs for taxonomy construction, using fine-tuning and prompting approaches. We also present a novel taxonomy construction dataset in the computer science domain. From a model-based perspective, we assess the consistency of generated taxonomies using structural constraints, while prior works emphasize performance compared to ground truth. 

\section{Conclusion}
\label{sec:conclusion}
This paper presents using LLMs in taxonomy construction, which involves creating hierarchical relations frequently used in software models. We conduct experiments on both a widely adopted hypernym dataset, along with a novel dataset derived from a computer science ontology. We incorporate the concept of \textit{consistency}, a widely employed principle in model-based software modeling, into taxonomy construction. We demonstrate that enhancing consistency leads to an improvement in the quality of the taxonomy compared to the ground truth.
 We compare fine-tuning and prompting approaches and show that the prompting method outperforms fine-tuning, especially when the training dataset is small. 
 
 This work discovers the ability of LLMs for taxonomy construction and we hope it can serve as a potential guideline for integrating LLMs with other taxonomy construction methods to further improve the quality of resulting taxonomies. This improvement may involve (1) using complex prompting methods \cite{yao2022react} to integrate other tools in the taxonomy construction process, (2) enhancing the performance of prompting methods through constraint-aware post-processing, (3) encoding domain-specific definition of the taxonomy relations, as well as (4) extending the comparison to other domains with various types of constraints.


 


\bibliographystyle{IEEEtran}
\bibliography{main}

\end{document}